# Agentic coding without the cloud: evaluating open-weight large language models on longitudinal data preparation tasks

Mack Nixon[1*], Liam Wright[2], Yevgeniya Kovalchuk[1], Alison Fang-Wei Wu[2], Martin Danka[2], Andy Boyd[3], David Bann[2*]

[1]Centre for Advanced Research Computing, University College London, London, UK

[2]Centre for Longitudinal Studies, University College London, London, UK

[3]UK Longitudinal Linkage Collaboration, University of Bristol, Bristol, UK

*Corresponding authors: Mack Nixon, Centre for Advanced Research Computing, University College London, London, UK. mack.nixon@ucl.ac.uk. David Bann, Centre for Longitudinal Studies, University College London, London, UK. david.bann@ucl.ac.uk

Large language models (LLMs) and agents are now widely used tools in code development, with data typically sent to third-party cloud-based models. Their adoption in research using personal data is constrained by governance requirements that typically prohibit data transmission to external services. Locally deployable open-weight models offer an alternative since sensitive data never leave the local environment. We introduce an open-source framework for evaluating the efficacy of AI agents powered by open-weight LLMs on one of the most persistent bottlenecks in research on longitudinal population studies: data preparation. The framework comprises: a curated ground-truth dataset (cleaning scripts preparing six sweeps of data from a British cohort study), task definitions encompassing tasks such as category harmonization and multi-wave merging, and automated routines for evaluating the LLM-produced R code and outputted data. We benchmark LLMs across the (consumer grade) deployment spectrum to assess their efficacy in 20 data preparation tasks (creation of 102 variables). Current state-of-the-art, 31-35B parameter models almost saturated our benchmark ('average task completion' up to 87.9%). The performance of open-weight LLMs running on consumer-grade hardware shows promise of a viable path toward AI-assisted data preparation in governance-restricted research settings. Our framework is publicly available at: https://github.com/UCL-ARC/RRBench.

## Introduction

In quantitative health and social sciences, considerable time is spent on repetitive data preparation. Tasks such as finding relevant variables, recoding survey responses, constructing composite indices, handling missingness, and merging datasets are essential but labor-intensive. Assuming a single study requires three months of data preparation, repeating this process across 1,000 projects equates to roughly 250 person-years of largely duplicated effort.[1] This cumulative burden creates a persistent bottleneck, slowing the pace at which data can be translated into scientific insight. Moreover, researchers frequently make coding mistakes, leading to inaccurate results; a recent study of major economics and political science journals identified coding errors in a quarter of papers assessed.[2]

Recent advances in generative artificial intelligence offer a promising way to mitigate these challenges. Large language models (LLMs) are increasingly capable of interpreting data dictionaries, transforming raw datasets, and generating analysis code from plain-language instructions.[3] When embedded within agentic frameworks, these models can also execute and iteratively refine such code (see Table 1 for a glossary of terms). Yet, the adoption of LLMs in survey and administrative data preparation is constrained by stringent data-governance requirements. Trusted research environments (TREs)—now the default for accessing sensitive data, such as electronic health records—allow data curation and analysis only within a secure environment, with access to the internet heavily restricted. But even less restrictive data-access agreements, allowing data to be stored on one's own machine often still prohibit transmitting individual-level data to external cloud services, such as frontier LLMs.[4]

**Table 1. Glossary of terms.**

|  |  |
|---|---|
| **Agentic AI** | A system that uses one or more AI agents. |
| **Agentic framework** | The application that sequences and orchestrates AI agents. |
| **AI agent** | An AI paradigm that connects LLMs to programmatic tools (e.g., web search) to enable LLMs to interact with their environment or do more than just token prediction. |
| **Benchmark** | A testing method commonly used in evaluating the performance or "intelligence" of LLMs. Benchmarks explore different notions of intelligence or aspects of performance, from general knowledge to critical thinking and domain-specific skills such as coding. |
| **Hallucination** | An assertion made by an LLM that is inaccurate, internally contradictory or fabricated. |
| **Large language model (LLM)** | A deep-learning model trained on large volumes of text data, generally designed to predict the next token (a unit of text) in a sequence. |
| **Lightweight** | When referencing frameworks (e.g., a 'lightweight script'), a small code footprint with few dependencies. When referencing models, fewer parameters and smaller model size. |
| **Native tool-calling capabilities** | Models trained or fine-tuned to be aware of how to directly use specific tools. |
| **Next-token prediction** | The pre-training task of an LLM, analogous to predicting the next word in a sentence. |
| **Parameters** | The trainable model weights of an LLM. |
| **Pre-training** | The initial phase of training an LLM on huge amounts of unlabeled data to learn and capture general patterns. |
| **Temperature** | A model hyperparameter (setting) that controls the randomness of an LLM's output. Values typically range between 0 and 1.5, with values closer to 0 ensuring greater determinism in results at the expense of model flexibility. Sometimes interpreted as 'creativity'. |
| **Token** | A unit of text interpreted and generated by an LLM. For example, "socioeconomic" may consist of two tokens: [socio] + [economic]. |
| **Tool** | A program or function an LLM has access to and can invoke to perform actions beyond text generation, such as searching the web or running code. |
| **Value labels** | Human readable labels for numeric codes. |

There are several ways of using LLMs with individual-level data, including: deploying open-weight models locally (i.e., data never leave the device they reside on); prompting LLMs hosted in public clouds but only exposing metadata or synthetic versions of datasets; or prompting LLMs hosted in approved secure cloud environments or institutional servers. Here, we focus on open-weight models deployed on consumer-grade hardware, permitting AI-assisted workflows on sensitive data. However, while such LLMs perform well on general software-engineering benchmarks (e.g., HumanEval, MBPP, APPS, and SWE-bench),[5–8] these benchmarks provide limited evidence on the domain-specific capabilities required for quantitative (survey) data preparation, where tasks include understanding (imperfectly) labelled metadata, correctly constructing derived measures (e.g., body mass index [BMI] in $kg/m^2$ or summary measures of mental health using accepted scales), and reconciling heterogeneous coding systems (e.g., harmonizing measures of employment status across multiple sweeps).

While some data cleaning benchmarks such as AutoDCWorkflow and DataSciBench exist,[9,10] they do not capture the multiple types of tasks noted above. Further, tests tend to focus on programming languages commonly used in computer science, such as Python,[11] rather than on languages used more frequently by quantitative researchers such as R.[12] In this paper, we address these gaps by introducing a framework for evaluating locally deployable LLMs on realistic data preparation challenges. Our contributions are threefold: (1) a curated benchmark that reflects the steps required to prepare (longitudinal) health and social science survey data for analysis; (2) a comparative assessment of model performance across open-weight LLMs of different sizes; and (3) an open-source pipeline enabling researchers to adapt the evaluation framework to their own datasets, data preparation tasks, and governance settings. Together, these resources seek to offer a practical foundation for integrating AI into locally secured data-preparation workflows.

## Methods

We developed an open-source framework, available at https://github.com/UCL-ARC/RRBench, to systematically evaluate LLMs on survey data preparation tasks. The framework comprises (1) a curated ground-truth dataset, (2) task definitions containing explicit data requirements and more general ('lite') data requirements, and (3) automated routines for generating, evaluating, and comparing LLM outputs. This setup is intended to facilitate benchmarking and comparison of LLM capabilities in data-centric task analysis. See Figure 1 for a visual overview.

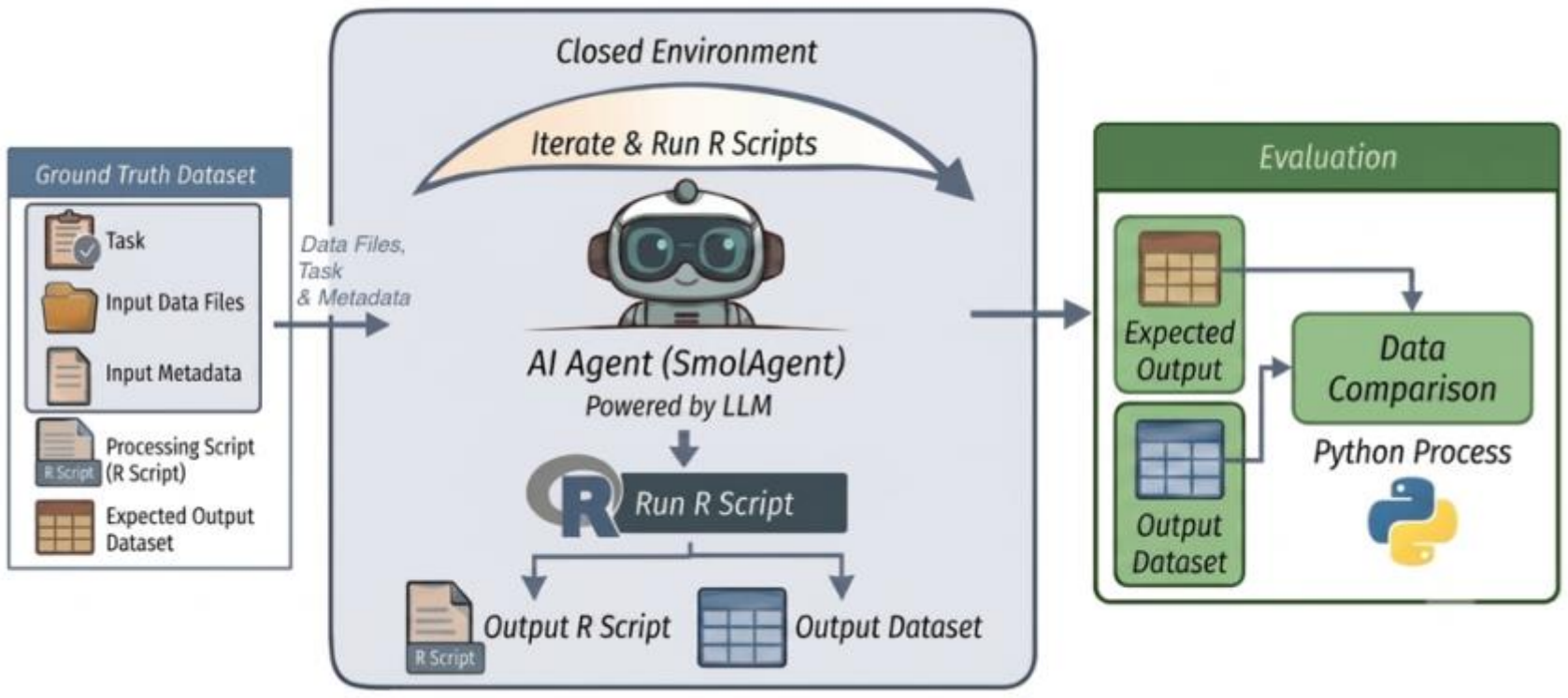

**Figure 1. Overview of the evaluation framework**. Each (human curated) ground-truth task comprises input data files, metadata, a processing script, and the expected output dataset produced by running the processing script on the input data files. The AI agent (SmolAgents) receives input files and metadata, then iteratively generates and executes an R script within a closed local environment. Agent performance is evaluated via data comparison, where an automated Python process compares the dataset output by an LLM against the ground-truth dataset.

### *Ground-truth data*

The dataset we used to construct our 'ground truth' for the benchmarking was the Next Steps Core dataset (https://github.com/cls-data/ns_core), a cleaned, research-ready version of data from Next Steps,[13] a longitudinal study of young people surveyed intermittently from age 14 years to (most recently) age 32 years, which has been used in hundreds of scientific studies.[14] The core dataset processing scripts were created by experienced cohort researchers and validated with independent code review. The dataset, which is stored in 'wide' format, contains multiple variables, some related to the young person (YP) and some to their parents, some fixed across time (e.g., sex, ethnicity) and some varying longitudinally (e.g., mental health), but most created by combining multiple variables from different survey sweeps (e.g., for ethnicity, using the first non-missing observation).

***Tasks***

The ground truth was divided into 20 distinct ‘tasks’, each of which comprised inputs available to the agent and expected outputs to evaluate performance. The inputs were (1) a prompt detailing the variable(s) to be made and general design decisions (e.g., variable naming conventions), (2) a set of input data files, and (3) metadata (in JSON format) for the raw variables and datasets (e.g., variable names and labels, and value labels). The outputs were an LLM-written R script and a dataset generated by the LLM upon executing the script. In addition, the same outputs for each corresponding task were produced by a human researcher (i.e., R scripts transforming raw Next Steps datasets into the Next Steps Core dataset and the corresponding dataset); these were not available to the LLM but used for benchmarking.

Each task consisted of preparing variables across commonly used domains in longitudinal data analysis such as demographic, education, health, and socioeconomic variables. Some tasks required the derivation of a single variable from information across several sweeps (e.g., a sex variable with minimal missingness across multiple sweeps); others required the creation of multiple age-specific variables (e.g., own and parental income, BMI, mental health). Each task required understanding of the source metadata (variable names, labels, and value labels) at varying levels of complexity (Figure 2). Tasks were categorized by their constituent operations through manual review (Table S1).

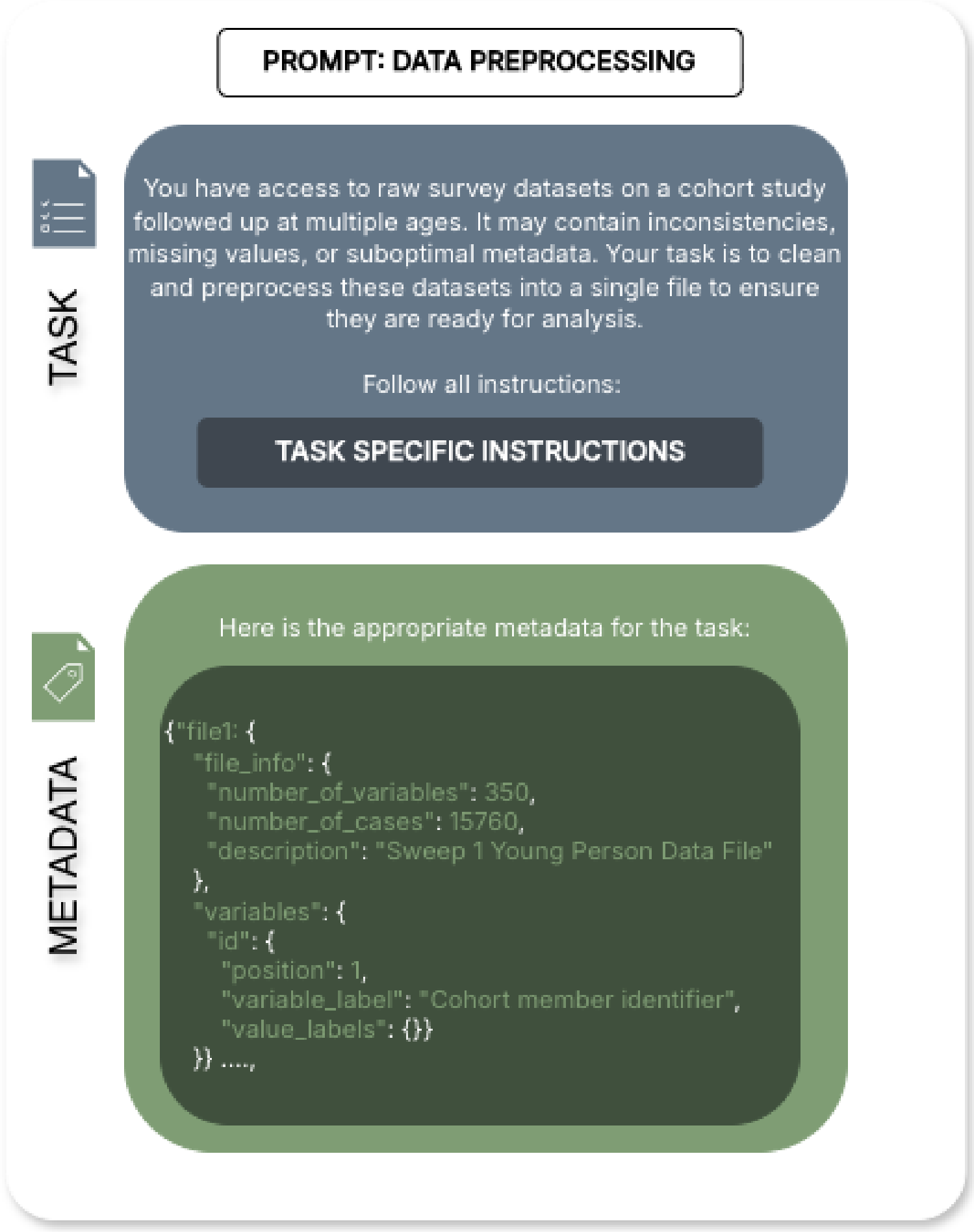

**Figure 2. Structure of the prompt provided to the AI agent**. The prompt consists of two components: a task description instructing the agent to clean and preprocess raw survey datasets into a single research-ready file, and structured metadata in JSON format describing the input data files, including variable names, labels, value labels, and file-level information such as the number of variables and cases.

In our primary analyses, detailed prompts were provided for each task to ensure that LLM evaluations were not biased due to misalignment with the decisions made in the ground-truth dataset; for each task, the detailed prompt specified code logic, required outcome variables, and response categories.

### *Agentic Workflow*

Each task was executed autonomously by an AI agent. The AI agent was powered by an underlying LLM that was configured with task-relevant context and a defined set of available tools. Many agentic frameworks employ varying orchestrations of LLM interactions, output parsing, context memory, and tool calling.[15] We opted for Hugging Face's SmolAgents[16] as it is lightweight and does not require an LLM with native tool-calling capabilities. This enabled the use of smaller open-weight models that could run on consumer-grade hardware. Each task was run a total of three times with a temperature of 0.8. Though this differs from the common recommendation of a lower temperature for coding tasks,[17] we observed that at lower temperatures, the agent would often remain stuck solving initial bugs in the R code it had produced.

### *Model Evaluation*

To evaluate the completion of each data preparation task, we systematically compared the generated output dataset with the expected, ground-truth dataset. We used a range of approaches to ensure we correctly matched outputs with their corresponding ground-truth datasets (see the accompanying GitHub repository).

### *Variable-level Comparisons with Ground-truth Datasets*

To quantify how well each model performed, we compared the output dataset from a given task with that produced by a human researcher. For each task, we calculated three measures of performance and then averaged these across tasks to get a balanced evaluation of a model's performance. The completeness metric measured the proportion of expected variables that were present in the LLM-processed output for a given task. The correctness metric measured the proportion of the recovered variables that met the prespecified agreement threshold when compared cell by cell with the corresponding ground-truth variable. A recovered variable (a variable in both ground-truth and LLM-generated datasets) was considered correct if a prespecified fraction of its cells exactly matched the ground truth; this fraction was set to 95% for categorical variables to allow for minor differences in variable categorization, e.g., where missing-value coding or category taxonomies differ across sweeps and a single harmonized scheme must be chosen. For continuous variables, the criterion was a normalized root mean squared error below $10^{-4}$ (effectively requiring exact reproduction, with a small tolerance to accommodate floating-point rounding). The balanced metric combined the completeness and correctness via multiplication into a single summary score; their product can be interpreted as the proportion of all ground-truth variables that were both attempted and correctly reproduced.

A task was considered fully completed only when all expected variables in the output dataset met the prespecified agreement threshold described above. We also calculated the mean task completion metric, defined as the average proportion of expected variables within each task that met the threshold. Finally, we manually reviewed a subset of generated processing scripts to help characterize common errors.

***Consistency with Ground-truth Dataset in Downstream Analyses***

A limitation of variable and task-level comparisons is that even small discrepancies (e.g., miscoded variable) can lead to large, material mistakes in final downstream results (e.g., regression coefficients from models using the cleaned data).[18] To evaluate the impact of LLM performance in data preparation on real-world downstream analyses, we conducted a series of analyses chosen to reflect realistic research questions of relevance to both health and social science (outcomes were BMI and income). We selected three groups of variables for which the LLM performance rates were relatively high, medium, and low, respectively. We then tested associations between the predictor variables and the outcomes using regression analyses and compared coefficients with those from the ground-truth dataset (see Results).

### *LLM Choice*

Models were selected to assess LLM performance across different model/parameter scales to inform deployment options (e.g., LLMs that could run on low- vs high-end consumer-grade hardware), model providers, and mixture-of-expert (MoE) vs dense architectures.[19]

Models were required to satisfy two criteria. First, the ability to handle context windows of at least 100,000 tokens, given that each task, including the SmolAgents tool-calling prompt template, required approximately 4,000 tokens on average (upper bound ~13,000 tokens), and the agent needed sufficient remaining context to explore the data, generate code, and iteratively correct errors. Second, to have been released within the prior 12 months (as of May 2026), so that models reflected the current state of the art. Table 2 summarizes the models evaluated and their approximate deployment requirements.

**Table 2. Details of the evaluated open-weight large language models (LLMs).**

| Model | Architecture | Total Params | Active Params | Deployment Tier |
|---|---|---|---|---|
| **Qwen3.6-35B-A3B**[28] | MoE | 35B | 3B | High-end consumer |
| **Qwen3.5-35B-A3B**[29] | MoE | 35B | 3B | High-end consumer |
| **Gemma 4 31B**[30] | Dense | 31B | 31B | High-end consumer |
| **Devstral-Small-2-24B-Instruct-2512**[31] | Dense | 24B | 24B | High-end consumer |
| **GPT-OSS-20B**[32] | MoE | 21B | 3.6B | Consumer |
| **Ministral-3-14B-Instruct-2512**[33] | Dense | 14B | 14B | Consumer |
| **Qwen3.5-9B**[29] | Dense | 9B | 9B | Consumer |
| **Gemma 4 E4B**[34] | Dense | 8B | 4.5B | Lightweight |

*Notes: High-end consumer: single 24GB GPU (e.g., RTX 4090/3090) or 32GB+ Mac; Consumer: mainstream 8–16GB GPU or standard laptop; Lightweight: any 4GB+ GPU, iGPU, or CPU-only. All deployment tiers are approximated from their respective quantized models. Mixture-of-experts (MoE) models must load all parameters into memory, even though only a fraction of parameters are active per token. All evaluated models are open-weight and released under the Apache 2.0 license, permitting use for research purposes.*

***Additional and sensitivity analyses***

To assess sensitivity to the choice of categorical accuracy thresholds, we measured LLM performance across a range of threshold values; we also measured mean concordance between variables to provide a single continuous measure of agreement. In addition to the detailed prompts used in main analyses, we also tested performance with a less detailed (‘lite’) prompt which contained brief instructions that specified only the required variables and response categories (e.g., ‘Derive one consolidated variable named sex containing information on the cohort member's sex’). Finally, since the ground truth required detailed coding instructions on handling of missing values (e.g., to distinguish refusals from inapplicables), we repeated evaluations excluding missing values to inform whether this impacted performance.

## Results

### *Variable-level Performance*

We evaluated eight LLMs on 20 data-preparation tasks drawn from the Next Steps longitudinal study. Results for the detailed prompt are presented in two stages: first, overall variable-level and task-level results across all tasks (Table 3); second, breakdown of performance by individual task (Table S1). Results for the 'lite' prompt are presented by overall variable-level and task-level aggregates across all tasks (Table S2).

Table 3 lists the aggregated performance results for each model. Gemma 4 31B scored highest (Balanced variable performance: 85.9%; Complete tasks: 75%), followed by Qwen3.6-35B-A3B (81.7%; 73.3%). Completeness was high across most models; even the worst overall performing model among those that produced valid outputs (Ministral-3-14B-Instruct-2512) created 95.4% of the ground-truth variables. The principal differentiator was the Correctness metric, with values ranging between 55.1% and 86.2%. The best-performing model completed 75% of tasks fully in a single agentic pass. At the lower end of the scale, Gemma 4 E4B failed to produce any successful output across all tasks and attempts. Inspection of the agent traces revealed that this was not due to deficient coding ability, but rather the model's inability to reliably produce responses in the structured format required by SmolAgents' tool-calling interface.

**Table 3. LLM performance under the accuracy threshold (≥0.95 categorical, ≤0.0001 numeric).**

| | Variable-level Metrics (pooled across tasks) | | | Task-level Metrics (computed per task, then averaged) | |
|---|---|---|---|---|---|
| **Model** | **Completeness (%)** | **Correctness (%)** | **Balanced Performance (%)** | **Complete Tasks (%)** | **Avg Task Completion (%)** |
| Gemma 4 31B | 99.7 | 86.2 | 85.9 | 75.0 | 87.9 |
| Qwen3.6-35B-A3B | 99.3 | 82.2 | 81.7 | 73.3 | 87.6 |
| GPT-OSS-20B | 98.0 | 79.3 | 77.8 | 60.0 | 81.0 |
| Devstral-Small-2-24B-Instruct-2512 | 97.1 | 76.4 | 74.2 | 60.0 | 79.7 |
| Qwen3.5-35B-A3B | 98.0 | 75.3 | 73.9 | 61.7 | 79.6 |
| Qwen3.5-9B | 98.7 | 62.9 | 62.1 | 53.3 | 68.4 |
| Ministral-3-14B-Instruct-2512 | 95.4 | 55.1 | 52.6 | 38.3 | 62.0 |

*Notes: Completeness (%): proportion of ground-truth variables produced by a model (e.g., 100% means all 102 variables were generated by a given model). Correctness (%): proportion of all the variables produced by a model that meet the accuracy threshold (≥95% cell-level agreement for categorical variables; normalized RMSE ≤$10^{-4}$ for continuous variables; analogous to precision;* 100% means every variable a given model has produced meets the threshold*). Balanced Performance (%) = Completeness × Correctness: proportion of ground-truth variables both produced and correct. Complete Tasks (%): share of all tasks, in which every expected variable met the accuracy threshold. Avg Task Completion (%): mean of the within-task proportion of variables meeting the accuracy threshold across all tasks. Variable-level metrics pool all variables across tasks into a single denominator; task-level metrics are computed per task and then averaged. Higher values indicate better performance. Gemma 4 E4B failed to produce any valid results and was omitted.*

Tasks involving concepts with common or near-universal definitions (e.g., sex, ethnicity, BMI and language) were consistently completed by most models, often achieving three out of three successful runs, even for smaller scale models (see Table S1). In contrast, tasks that depend heavily on survey-specific metadata—particularly those requiring the model to interpret bespoke variable naming conventions, navigate complex coding frames, or follow domain-specific derivation logic—proved substantially harder. Housing tenure was not fully completed by any of the models in any run, while Education Aim (current qualification studied), Education (YP),

Economic Activity (Parents), NS-SEC (YP; occupational/socio-economic classification) and Household Income were completed less than 30% of the time on average across all runs.

Figure 3 and Table S1 disaggregates LLM performance by the type of data-preparation operation. Wave-specific logic, taxonomy compression, and multi-input derived variable construction emerged as the most challenging operation categories.

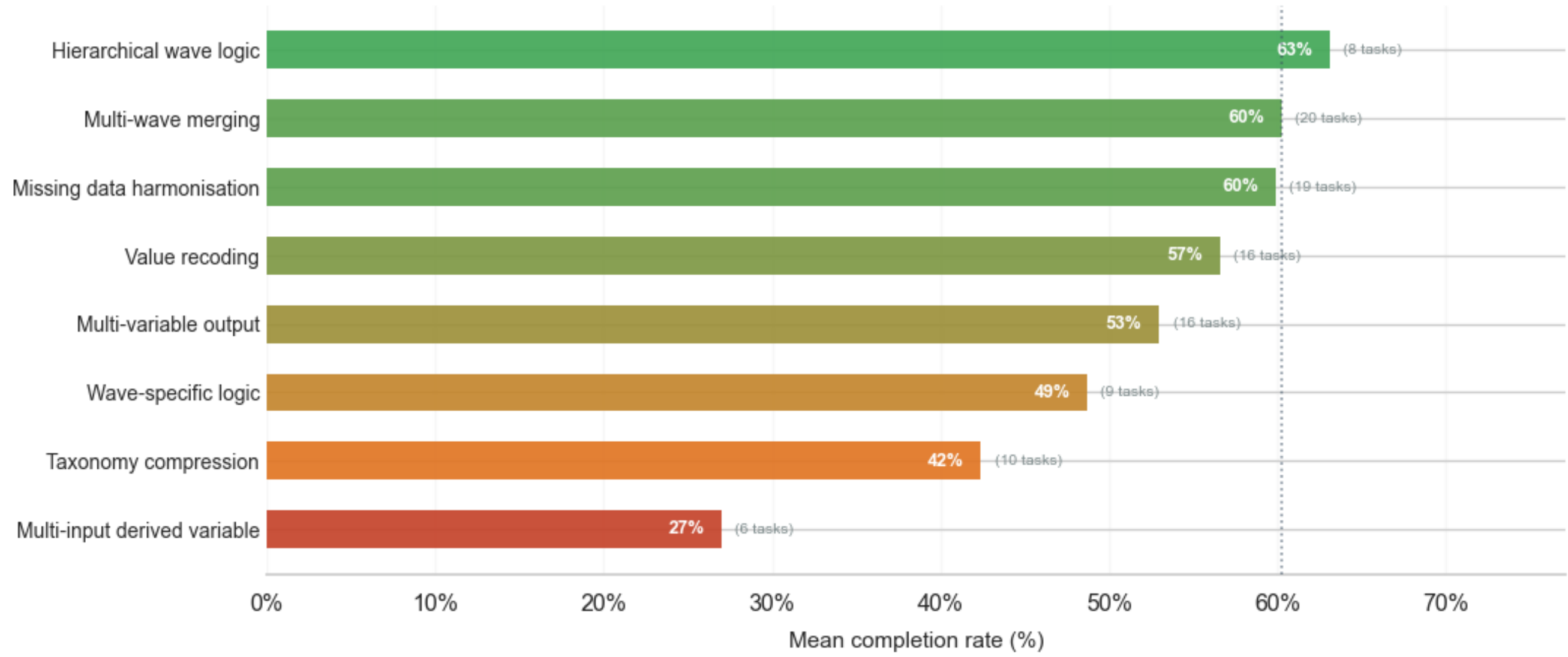

**Figure 3. Task complexity index, inferred from average task completion rate.** For each data-preparation operation, we measured the average completion rate for all tasks containing that operation.

Figure 4 shows the relationship between the task complexity and task success. Three proxies for complexity were considered: the length of the task prompt (which largely reflects the volume of metadata), the length of the ground-truth R script, and the number of distinct data-preparation operations involved. Individually, all three proxies showed a negative association with task success: longer prompts (Pearson r: -0.75), longer scripts (Pearson r: -0.64), and a greater number of operations (Pearson r: -0.60), each corresponded to lower completion rates.

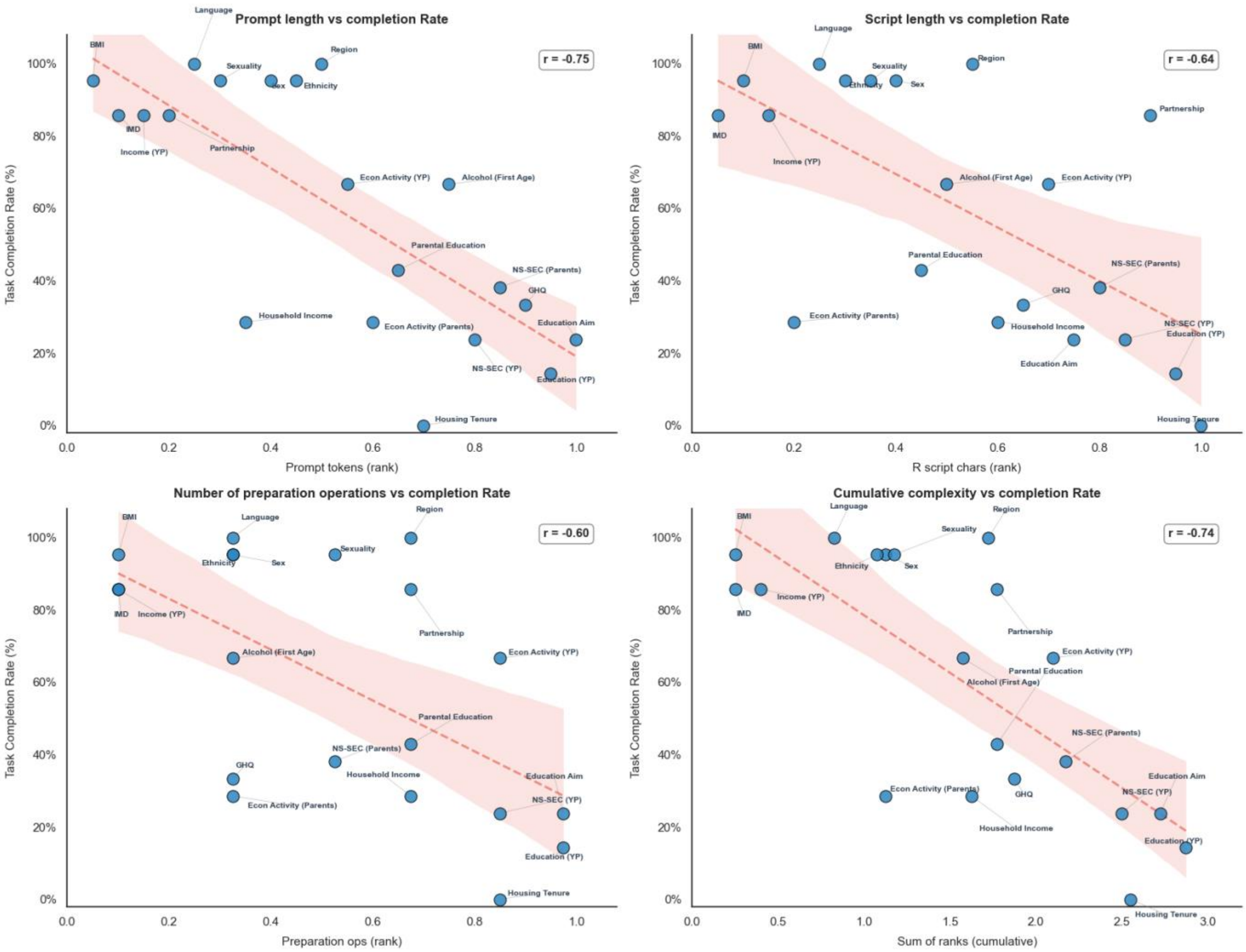

**Figure 4. Task complexity plotted against task success.** An individual task consists of the task tokens, metadata tokens and the SmolAgents prompt template tokens, all of which are added when calculating the task prompt length. Each of the ‘R script length’, ‘Prompt length’, and ‘Number of wrangling operations’ measurements has been normalized using their respective rank (e.g., a task's shortest prompt received rank 1, longest received rank N and each then normalized between 0 and 1).

*Performance for Downstream Analyses*

To illustrate how variable-level errors propagate into substantive findings, we re-ran three simple bivariate regressions using datasets produced by the best- and worst-performing models (Table 4). The datasets produced by Gemma 4 31B (the best-performing LLM) yielded coefficients identical to the ground truth across all six regressions. For derivation of ethnicity, for example, Gemma 4 31B harmonized using category labels from earlier sweeps rather than the more general labels from later sweeps as chosen in the ground truth (Figure S1).

**Table 4. Linear regression with a single binary exposure variable using ground-truth datasets and LLM generated datasets.**

| | **Ethnicity (White vs Black African)** | **Economic activity at age 19 (in paid work vs other)** | **Sex (male vs female)** |
|---|---|---|---|
| ***Outcome: BMI*** | | | |
| **Ground truth** | −1.29 (95% CI: −2.32, −0.25) | 0.75 (95% CI: 0.34, 1.16) | −0.68 (95% CI: −1.02, −0.33) |
| **Gemma 4 31B** | −1.29 (95% CI: −2.32, −0.25) | 0.75 (95% CI: 0.34, 1.16) | −0.68 (95% CI: −1.02, −0.33) |
| **Ministral 3 14B** | −1.29 (95% CI: −2.32, −0.25) | 0.75 (95% CI: 0.34, 1.16) | −0.66 (95% CI: −1.01, −0.31) |
| ***Outcome: YP income*** | | | |
| **Ground truth** | 54.2 (95% CI: −6.04, 114) | −38.6 (95% CI: −64.2, −12.2) | 46.1 (95% CI: 25.4, 66.8) |
| **Gemma 4 31B** | 54.2 (95% CI: −6.04, 114) | −38.6 (95% CI: −64.2, −12.2) | 46.1 (95% CI: 25.4, 66.8) |
| **Ministral 3 14B** | 165 (95% CI: 106, 224) | 10.6 (95% CI: −14.8, 36.0) | 38.0 (95% CI: 17.1, 58.9) |

*Notes: Each regression was conducted using ordinary least squares with a single binary exposure variable. Three datasets were chosen (ground-truth and two LLM-generated datasets), and each outcome and exposure variable was constructed identically, except for small adjustments to account for variation in category names (e.g., for YP income, “25-50” vs “25 to 50” the same numeric midpoint was chosen). No covariates were included; each model is unadjusted. We report the exposure regression coefficients of each model.*

Ministral 3 14B (the worst-performing LLM) produced data that looked plausible at first glance but contained silent errors: the scripts ran without warnings, the variables appeared well-formed, and only cell-by-cell comparison against the ground truth revealed the problems. For income (age 32), Ministral 3 14B incorrectly labelled a set of NA values as “1”, instead of the correct “-1” (Figure S2). In the model predicting income at age 32 (£), this recoding bug caused large downstream effects e.g., increasing the ethnicity coefficient by £110.83 and reversing the direction of the economic activity coefficient (from -£38.64 to £10.60). When deriving sex, Ministral 3 14B used inaccurate handling of NA values, imputing NAs before harmonization instead of imputing NAs in a wave-specific order, leading to an inaccurately derived variable (Figure S3). However, since males and females were generally not misclassified, the direction of the association with BMI was preserved.

### *Additional and sensitivity analyses*

Performance was generally similar at thresholds ranging 80-95%, with steep declines at a 100% threshold (Figure S4). Mean concordance with the ground truth was very high (ranging 85.5% to 95.6%; Figure S5). When using a less detailed ‘lite’ prompt, mean concordance with the ground truth was generally high (86.9%, range 82.2%-90.2%; Figure S5). When assessed in terms of completeness and correct tasks, performance was lower than with the main prompts (e.g., a maximum of 96.4% completion and 63.1% correctness for GPT-OSS-20B, the highest performing model; Table S2). Correctness was considerably higher when excluding missing data from the comparisons (76.5% for Gemma 4 31B using the 'lite' prompt, Table S3; 94.4% using the main prompt, Table S4), suggesting that lower performance was driven by differences in missing data coding.

## Discussion

We evaluated eight open-weight LLMs, deployed entirely on local infrastructure, on 20 data-preparation tasks drawn from a British longitudinal cohort study. The best-performing model (Gemma 4 31B) accurately completed 75% of the tasks, with 85.9% of 102 variables being correctly derived. LLM performance generally rose with the number of model parameters and varied by task type: variables with near-universal definitions well represented in training corpora (e.g., sex, ethnicity, BMI) were reliably reproduced by most models, whereas tasks requiring understanding of survey-specific metadata, combined with logically complex derivation (e.g., housing tenure, household income) proved more difficult, even with derivation logic spelled out (in natural language) in the prompt. The principal differentiator was correctness rather than completeness: 'weaker' models reliably identified the variables to produce but were prone to hallucinating plausible yet incorrect preprocessing steps, such as recoding logic that did not align with the metadata. In downstream regression analyses, datasets produced by the best-performing model yielded coefficients identical to the ground truth, while a weaker model produced 'silent' errors: plausible-looking variables containing data quality issues that materially shifted the magnitude of regression estimates.

We extend the existing software-engineering benchmarks (e.g., SWE-bench, HumanEval, MBPP, and APPS[5–8]) and data cleaning benchmarks (e.g., AutoDCWorkflow and DataSciBench[9,10]) by capturing the domain-specific reasoning required for preparing research-ready longitudinal health and social science study data using R—the programming language more widely used in health and social sciences rather than Python which is targeted by existing benchmarks.[12] Our benchmark is built from human-curated, independently reviewed, ground-truth data derived from diverse tasks (e.g., multi-wave harmonization, hierarchical derivation logic, and reconciliation of heterogeneous coding frames). This provides a novel approach for comparing tabular data similarity. Our framework is designed to be portable: the structure (ground-truth input data, task definitions with metadata, expected output datasets, and automated comparison routines) can be readily adapted to other tabular data transformation tasks. Extending the benchmark to additional cohorts or different programming languages would both validate the generalizability of our findings and create a broader evidence base for model selection. The open-source release of our framework (https://github.com/UCL-ARC/RRBench) is intended to facilitate exactly this kind of community-driven expansion.

The evaluation framework presented in this paper deliberately measures LLMs' fully autonomous performance: each model received a task prompt and was required to produce a complete, research-ready dataset without human intervention, which represents the most demanding possible test of LLM capability. While this constraint was necessary to evaluate models here, researchers are unlikely to use these models as fully autonomous pipelines in practice. The more immediate use case is using local LLMs as coding copilots accessed within IDEs: producing first-pass preparation scripts with 'lite' prompts, which the researcher then reviews, validates, corrects, or queries. This is particularly advisable when using smaller models, where silent errors can materially affect downstream results, reinforcing the need for human-in-the-loop review. Even with human oversight, partial productivity improvements scaled across the many thousands of population datasets held worldwide could save years of research time,[20] which could be re-allocated to, for example, conduct more ambitious multi-dataset research.[21,22]

Several patterns in our results point to potential limitations that warrant further investigation. Housing tenure was not fully completed by any of the evaluated models in any run, and tasks involving complex survey-specific metadata (e.g., household income, economic activity) proved consistently difficult. These failures share a common feature: they require the model not only to write syntactically correct code but to interpret intricate survey metadata, to combine that with relevant domain-specific information and translate that comprehension into logically complex derivation sequences. While a variable such as BMI has a near-universal definition well-represented in training corpora, housing tenure categories are survey-specific constructs that the model must learn entirely from the metadata provided in the prompt. Consistent with this, the prompt length showed the strongest negative association with task success. The prompt length was largely a function of the metadata embedded in the prompt, acting as a proxy for data complexity, although poorer performance may also be influenced by long-text degradation.[23,24]

Finally, LLM performance is sensitive to prompt design, and we acknowledge the substantial researcher degrees of freedom this introduces. This is demonstrated in the drop in performance when we provided fewer explicit derivation instructions. While we sought to optimize our prompts to maximize performance, the space of plausible prompt alternatives is vast, and we could not feasibly test all variants. Choices around phrasing, structure, ordering, and the level of explicit guidance provided can all plausibly affect the model performance, and different choices could have altered the results. Adding further complication is the evolving nature of prompting best practice, which may, in turn, differ by model. Our findings should therefore be interpreted as indicative of what is achievable with a reasonable prompting strategy rather than as a definitive ceiling on model capability. To help inform readers' own prompts, we include notes on prompt design in the GitHub repository. Future work could systematically test whether and how prompts could be improved further (e.g., via automated prompt optimization) while balancing performance with generalizability across different sets of LLMs.

Our evaluation used a single agent responsible for interpreting metadata, writing code, executing it, and debugging errors—a workload that places substantial demands on the underlying model. Shifting from single-agent to multi-agent architectures by decomposing the framework into specialized agents with narrower responsibilities could meaningfully improve performance,[25] albeit increasing complexity. A planner agent might first produce a structured derivation plan from the metadata; a coder agent could then translate the plan into an R script; a reviewer agent could validate outputs against metadata constraints and flag suspicious patterns (for instance, implausible category distributions or unexpected missingness) before returning results to the researcher. This all hinges on the readability and consistency of the metadata. Following standards such as the FAIR principles[26] can help ease the integration of LLMs with survey data.

Our evaluation framework could also be extended to incorporate the use of ‘LLM as a judge’ evaluation to increase the number of evaluation dimensions from data accuracy to script readability, reusability, and robustness.[27] Finally, future work could also examine the performance of local LLMs in other scripting languages, both those that are likely to be better represented in the training corpora (e.g., Python) and those that are less well-represented but widely used by social and healthcare researchers (e.g., Stata, SAS).

## Conclusion

Open-weight LLMs deployed entirely on local infrastructure can reproduce human-curated data preparation scripts for longitudinal population studies with meaningful fidelity; as with code generated by human developers or cloud-based models, human code review is still recommended. Given their already high performance, optimizing metadata to aid LLM understanding may offer larger performance gains than model-side improvements alone. The most performant models are still restricted to those deployable on high-end consumer-grade hardware. This highlights a need for investment in computing infrastructure that can improve accessibility to these models. Crucially, this infrastructure investment is qualitatively different from subscribing to cloud-based LLM services; open-weight models running on institutional hardware involve no transmission of sensitive data to external services, no ongoing per-query costs, and no dependency on commercial providers' data handling policies. For data holders managing sensitive data, or infrastructure that hosts them (e.g., TREs), these models could be deployed within secure perimeters, making AI-assisted data preparation available to researchers working under the most restrictive governance frameworks.

## Statements

### *Data availability*

Next Steps (NS) data can be accessed via an End User Licence Agreement from the UK Data Service (http://doi.org/10.5255/UKDA-SN-5545-12). Code to produce the 'ground-truth' NS Core dataset can be viewed on GitHub (https://github.com/CLS-Data/ns_core). The framework used in this paper is accessible at https://github.com/UCL-ARC/RRBench.

### *Acknowledgments*

The authors acknowledge the use of the UCL Unified AI Services Computing Facility (UAIServices@UCL), and associated support services, in the completion of this work.

### *Author contributions*

MN wrote the first draft and led analysis. MN, DB, and LW led the study design and contributed to analysis. All authors contributed to generating ideas, compiling material, reviewing and revising the text, as well as the accompanying repository. AW created the ground-truth dataset, with MD conducting code review and validation of the resulting outputs. Authors AB, DB and LW secured the funding.

### *Conflicts of interest*

None declared.

### *Funding*

DB and LW are funded by the Economic and Social Research Council (ES/W013142/1 and UKRI3344); and DB, LW and MN by the UKRI Digital Research Infrastructure (DRI) Programme (UKRI/ST/B000295/1). AB is funded by UKRI, including the Medical Research Council (MR/X021556/1) and the Economic and Social Research Council (ES/X000567/1). MD is funded by the ESRC-BBSRC Social-Biological (Soc-B) Centre for Doctoral Training (ES/P000347/1).

### *Use of Artificial intelligence (AI) tools*

LLMs such as ChatGPT and Claude were used to assist in code creation and revision of the text. Human authors drafted all sections and carefully edited/reviewed any suggestions made by the LLMs.

**Supplementary information for:**

## Agentic coding without the cloud: evaluating open-weight large language models on longitudinal data preparation tasks

Mack Nixon[1*], Liam Wright[2], Yevgeniya Kovalchuk[1], Alison Fang-Wei Wu[2], Martin Danka[2], Andy Boyd[3], David Bann[2*]

[1]Centre for Advanced Research Computing, University College London, London, UK

[2]Centre for Longitudinal Studies, University College London, London, UK

[3]UK Longitudinal Linkage Collaboration, University of Bristol, Bristol, UK

**Supplementary Table 1. Individual-task performance by selected LLMs (a) and constituent data-preparation operations (b), under the accuracy threshold (≥ 0.95 categorical, ≤ 0.0001 numeric)**

| Task | (a) Runs fully completed (of 3) | | | | | | | | (b) Data-preparation operations | | | | | | | | |
|---|---|---|---|---|---|---|---|---|---|---|---|---|---|---|---|---|---|
| | Gemma 4 31B | Qwen3.6-35B | GPT-OSS-20B | Devstral-Small-2-24B | Qwen3.5-35B | Qwen3.5-9B | Ministral 3 14B | Avg Completion % | MW | MH | HL | VR | TC | WL | MV | MI | Total |
| Language | 3 | 3 | 3 | 3 | 3 | 3 | 3 | **100** | ✓ | ✓ | ✓ | ✓ | | | | | 4 |
| Region | 3 | 3 | 3 | 3 | 3 | 3 | 3 | **100** | ✓ | ✓ | | ✓ | ✓ | ✓ | ✓ | | 6 |
| Sex | 3 | 3 | 3 | 3 | 3 | 3 | 2 | **95.2** | ✓ | ✓ | ✓ | ✓ | | | | | 4 |
| Ethnicity | 3 | 3 | 3 | 3 | 3 | 2 | 3 | **95.2** | ✓ | ✓ | ✓ | ✓ | | | | | 4 |
| Sexuality | 3 | 3 | 3 | 3 | 3 | 3 | 2 | **95.2** | ✓ | ✓ | | ✓ | | ✓ | ✓ | | 5 |
| BMI | 3 | 3 | 3 | 3 | 2 | 3 | 3 | **95.2** | ✓ | ✓ | | | | | ✓ | | 3 |
| Partnership | 3 | 3 | 3 | 3 | 3 | 3 | 0 | **85.7** | ✓ | ✓ | | ✓ | ✓ | ✓ | ✓ | | 6 |
| Income (YP) | 3 | 3 | 3 | 3 | 2 | 2 | 2 | **85.7** | ✓ | ✓ | | | | | ✓ | | 3 |
| IMD | 2 | 3 | 3 | 3 | 3 | 2 | 2 | **85.7** | ✓ | ✓ | | | | | ✓ | | 3 |
| Economic activity (YP) | 3 | 0 | 3 | 3 | 3 | 2 | 0 | **66.7** | ✓ | ✓ | ✓ | ✓ | ✓ | ✓ | ✓ | | 7 |
| Alcohol – first age derived | 3 | 3 | 1 | 1 | 2 | 2 | 2 | **66.7** | ✓ | | ✓ | ✓ | | | | ✓ | 4 |
| Parental education | 1 | 3 | 2 | 0 | 2 | 0 | 1 | **42.9** | ✓ | ✓ | ✓ | ✓ | ✓ | | ✓ | | 6 |
| NS-SEC (parents) | 3 | 3 | 2 | 0 | 0 | 0 | 0 | **38.1** | ✓ | ✓ | | ✓ | ✓ | | ✓ | | 5 |
| GHQ | 2 | 0 | 0 | 1 | 2 | 2 | 0 | **33.3** | ✓ | ✓ | | | | | ✓ | ✓ | 4 |
| Economic activity (parents) | 3 | 2 | 0 | 0 | 0 | 1 | 0 | **28.6** | ✓ | ✓ | | ✓ | | | ✓ | | 4 |
| Household income | 3 | 1 | 0 | 2 | 0 | 0 | 0 | **28.6** | ✓ | ✓ | | ✓ | ✓ | ✓ | ✓ | | 6 |
| Education aim | 1 | 1 | 0 | 2 | 1 | 0 | 0 | **23.8** | ✓ | ✓ | ✓ | ✓ | ✓ | ✓ | ✓ | ✓ | 8 |
| NS-SEC (YP) | 0 | 2 | 0 | 0 | 2 | 1 | 0 | **23.8** | ✓ | ✓ | | ✓ | ✓ | ✓ | ✓ | ✓ | 7 |
| Education (YP) | 0 | 2 | 1 | 0 | 0 | 0 | 0 | **14.3** | ✓ | ✓ | ✓ | ✓ | ✓ | ✓ | ✓ | ✓ | 8 |
| Housing tenure | 0 | 0 | 0 | 0 | 0 | 0 | 0 | **0** | ✓ | ✓ | | ✓ | ✓ | ✓ | ✓ | ✓ | 7 |
| ***Total*** | | | | | | | | | **20** | **19** | **8** | **16** | **10** | **9** | **16** | **6** | |

*Notes: Panel (a): each entry reports the number of times a given task was fully completed within three distinct attempts; Avg Completion % is the mean completion rate across all models and runs. Panel (b): ✓ indicates that the ground-truth script for the task involves the given operation. Gemma 4 E4B failed to produce any valid results and was omitted. Operations: (MW) multi-wave merging: joining data from different survey sweeps; (MH) missing data harmonization: standardizing missing-value codes; (HL) hierarchical wave logic: deriving values based on a priority order from sweeps; (VR) value recoding: remapping raw survey codes into new values; (TC) taxonomy compression: reducing categories according to a known classification; (WL) wave-specific logic: applying different transformation logic per wave; (MV) multi-variable output: producing more than one output variable; (MI) multi-input-derived variable construction: computing a variable from multiple inputs.*

**Supplementary Table 2. LLM performance for 'lite' prompts under the accuracy threshold (≥0.95 categorical, ≤0.0001 numeric)**

| | Variable-level Metrics (pooled across tasks) | | | Task-level Metrics (computed per task, then averaged) | |
|---|---|---|---|---|---|
| **Model** | **Completeness (%)** | **Correctness (%)** | **Balanced Performance (%)** | **Complete Tasks (%)** | **Avg Task Completion (%)** |
| GPT-OSS-20B | 96.4 | 63.1 | 60.8 | 50.0 | 67.5 |
| Gemma 4 31B | 97.4 | 63.1 | 61.4 | 48.3 | 67.4 |
| Devstral-Small-2-24B-Instruct-2512 | 94.1 | 59.0 | 55.6 | 45.0 | 65.2 |
| Qwen3.6-35B-A3B | 97.4 | 54.4 | 52.9 | 50.0 | 63.4 |
| Qwen3.5-35B-A3B | 95.8 | 54.6 | 52.3 | 43.3 | 61.5 |
| Qwen3.5-9B | 94.8 | 51.4 | 48.7 | 40.0 | 57.2 |
| Ministral-3-14B-Instruct-2512 | 85.9 | 41.8 | 35.9 | 36.7 | 49.0 |

*Notes: Same metric definitions as in Table 3. Gemma 4 E4B failed to produce any valid results and was omitted.*

**Supplementary Table 3: LLM performance for 'lite' prompts under the accuracy threshold (≥0.95 categorical, ≤0.0001 numeric) without missing values**

| | Variable-level Metrics (pooled across tasks) | | | Task-level Metrics (computed per task, then averaged) | |
|---|---|---|---|---|---|
| **Model** | **Completeness (%)** | **Correctness (%)** | **Balanced Performance (%)** | **Complete Tasks (%)** | **Avg Task Completion (%)** |
| Gemma 4 31B | 97.4 | 76.5 | 74.5 | 58.3 | 78.5 |
| Qwen3.6-35B-A3B | 97.4 | 68.1 | 66.3 | 56.7 | 77.4 |
| GPT-OSS-20B | 96.4 | 73.9 | 71.2 | 53.3 | 75.6 |
| Qwen3.5-35B-A3B | 95.8 | 66.2 | 63.4 | 51.7 | 72.5 |
| Devstral-Small-2-24B-Instruct-2512 | 94.1 | 67.7 | 63.7 | 50.0 | 72.1 |
| Qwen3.5-9B | 94.8 | 61.4 | 58.2 | 50.0 | 69.2 |
| Ministral-3-14B-Instruct-2512 | 85.9 | 60.8 | 52.3 | 41.7 | 60.9 |

*Notes: Same metric definitions as in Table 3. Gemma 4 E4B failed to produce any valid results and was omitted.*

**Supplementary Table 4: LLM performance under the accuracy threshold (≥0.95 categorical, ≤0.0001 numeric) without missing values**

| | Variable-level Metrics (pooled across tasks) | | | Task-level Metrics (computed per task, then averaged) | |
|---|---|---|---|---|---|
| **Model** | **Completeness (%)** | **Correctness (%)** | **Balanced Performance (%)** | **Complete Tasks (%)** | **Avg Task Completion (%)** |
| Qwen3.5-35B-A3B | 98.0 | 93.0 | 91.2 | 88.3 | 93.8 |
| Gemma 4 31B | 99.7 | 94.4 | 94.1 | 86.7 | 93.6 |
| Qwen3.6-35B-A3B | 99.3 | 86.5 | 85.9 | 78.3 | 90.2 |
| Devstral-Small-2-24B-Instruct-2512 | 97.1 | 90.9 | 88.2 | 75.0 | 89.1 |
| GPT-OSS-20B | 98.0 | 89.0 | 87.3 | 66.7 | 88.1 |
| Qwen3.5-9B | 98.7 | 79.8 | 78.8 | 66.7 | 81.3 |
| Ministral-3-14B-Instruct-2512 | 95.4 | 80.5 | 76.8 | 63.3 | 80.5 |

*Notes: Same metric definitions as in Table 3. Gemma 4 E4B failed to produce any valid results and was omitted.*

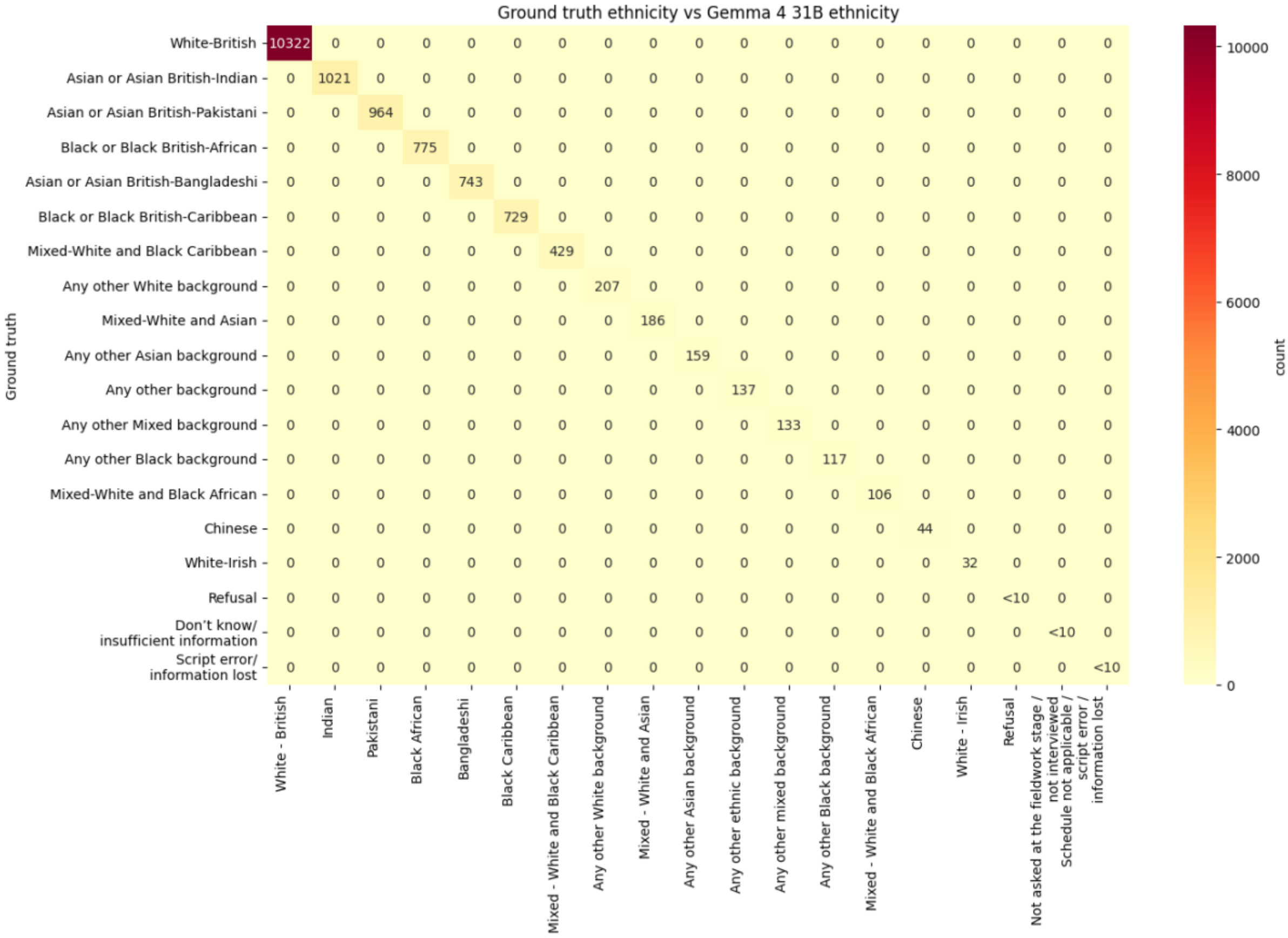

**Supplementary Figure 1. Crosstab comparing ethnicity data generated by Gemma 4 31B with those from the ground truth.** Values in the diagonal show full concordance, with minor differences in the naming of the value labels resolved via the matching algorithm.

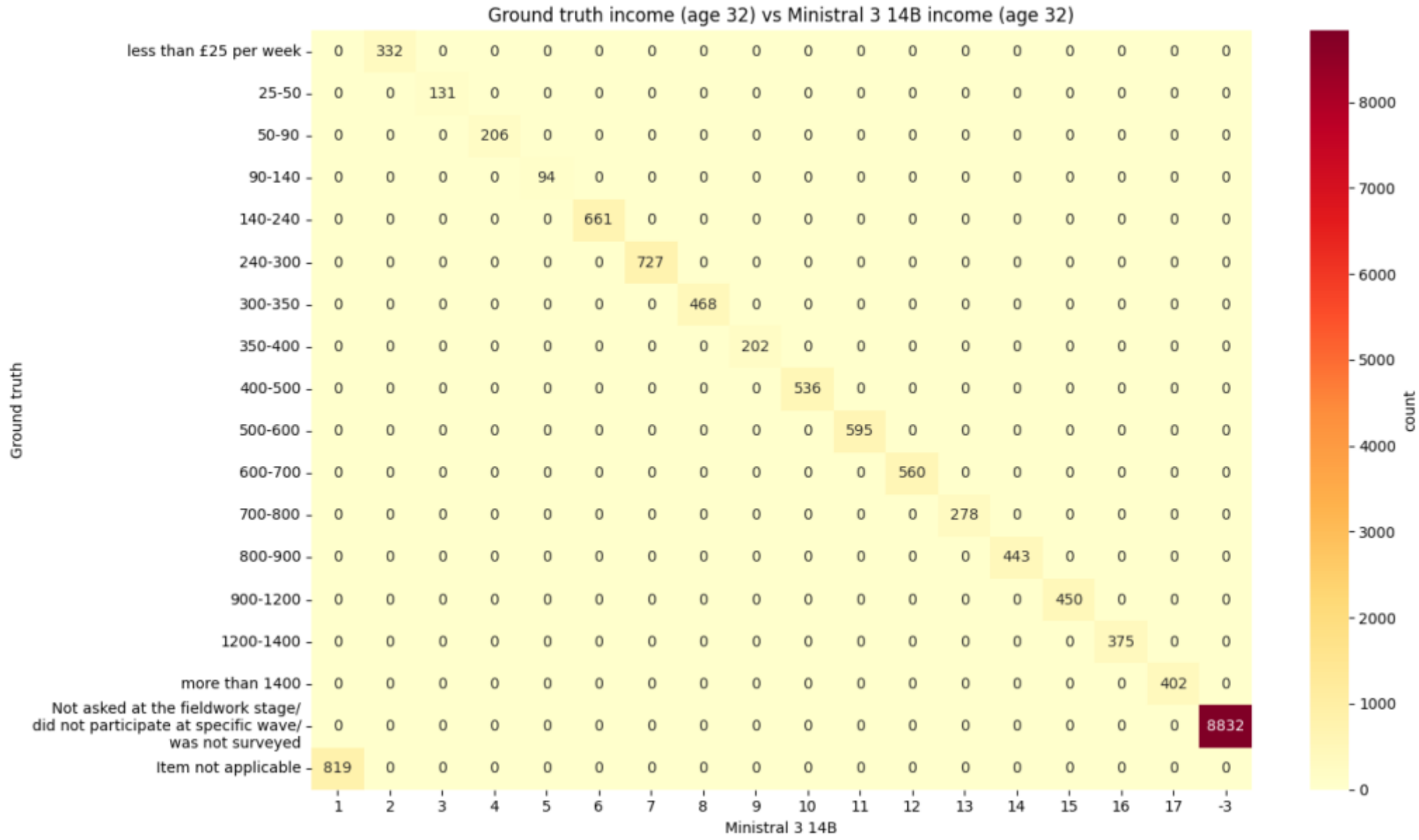

**Supplementary Figure 2. Crosstab comparing income data generated by Ministral 3 14B and those from the ground truth.** Values in the diagonal show concordance, despite differences in the naming of the value labels. 'Item not applicable' has been incorrectly labelled as a non-null value.

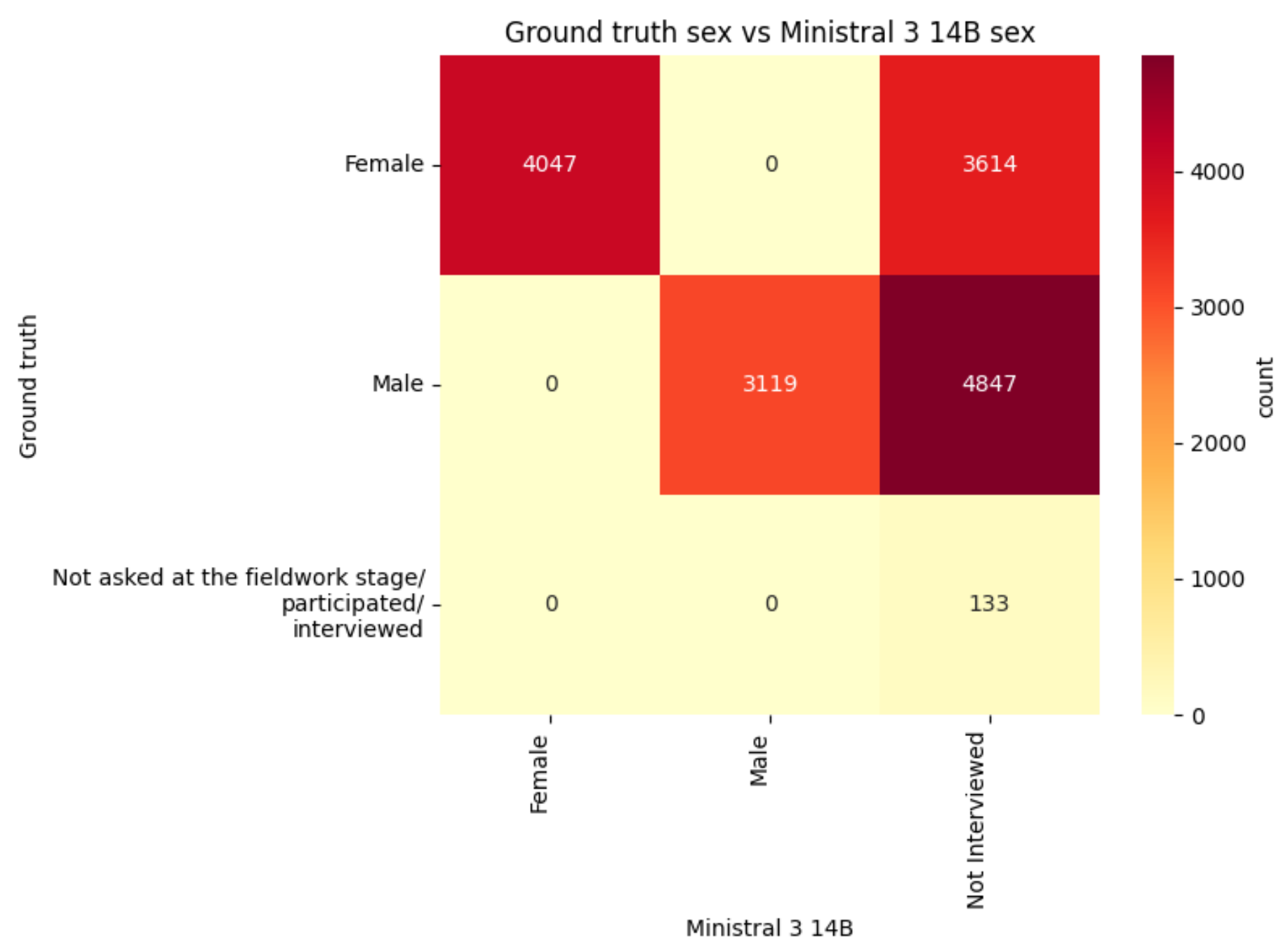

**Supplementary Figure 3. Crosstab comparing sex data generated by Ministral 3 14B and those from the ground truth.** Values in the diagonal show concordance. Ministral 3 14B incorrectly generated missing values that were not present in the ground-truth generated data.

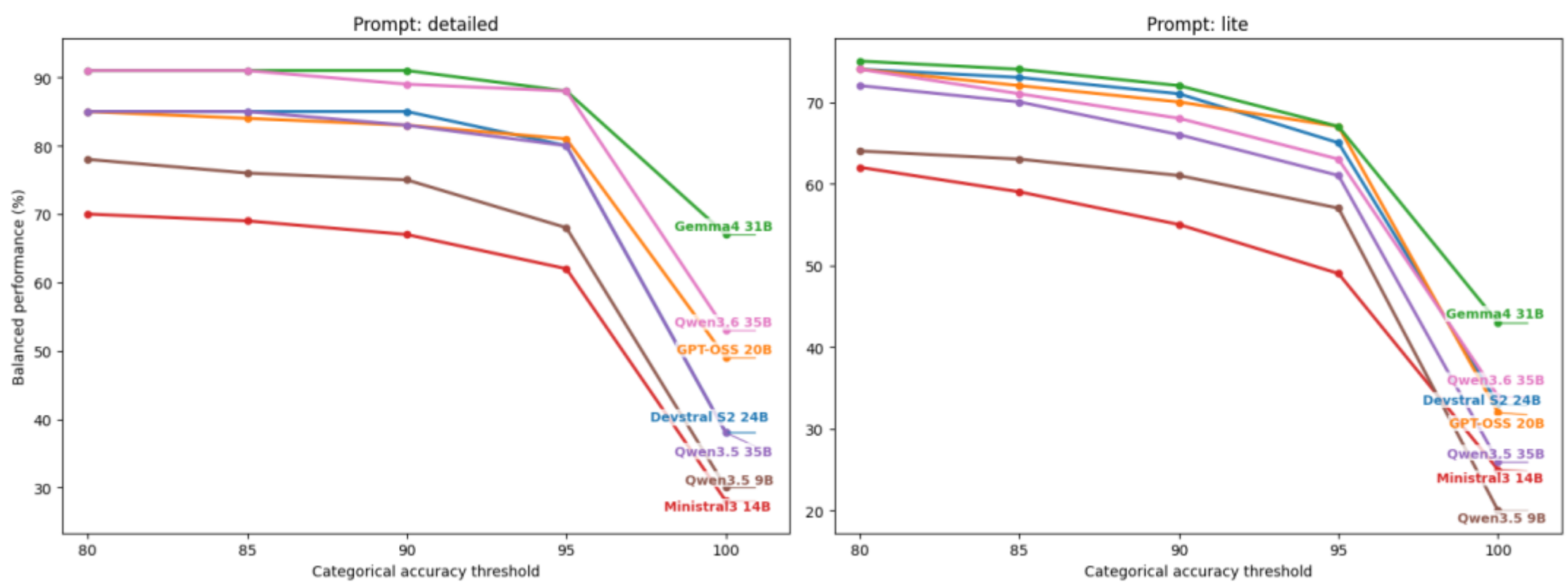

**Supplementary Figure 4:** Sensitivity analysis of 'balanced performance' with varying categorical accuracy threshold.

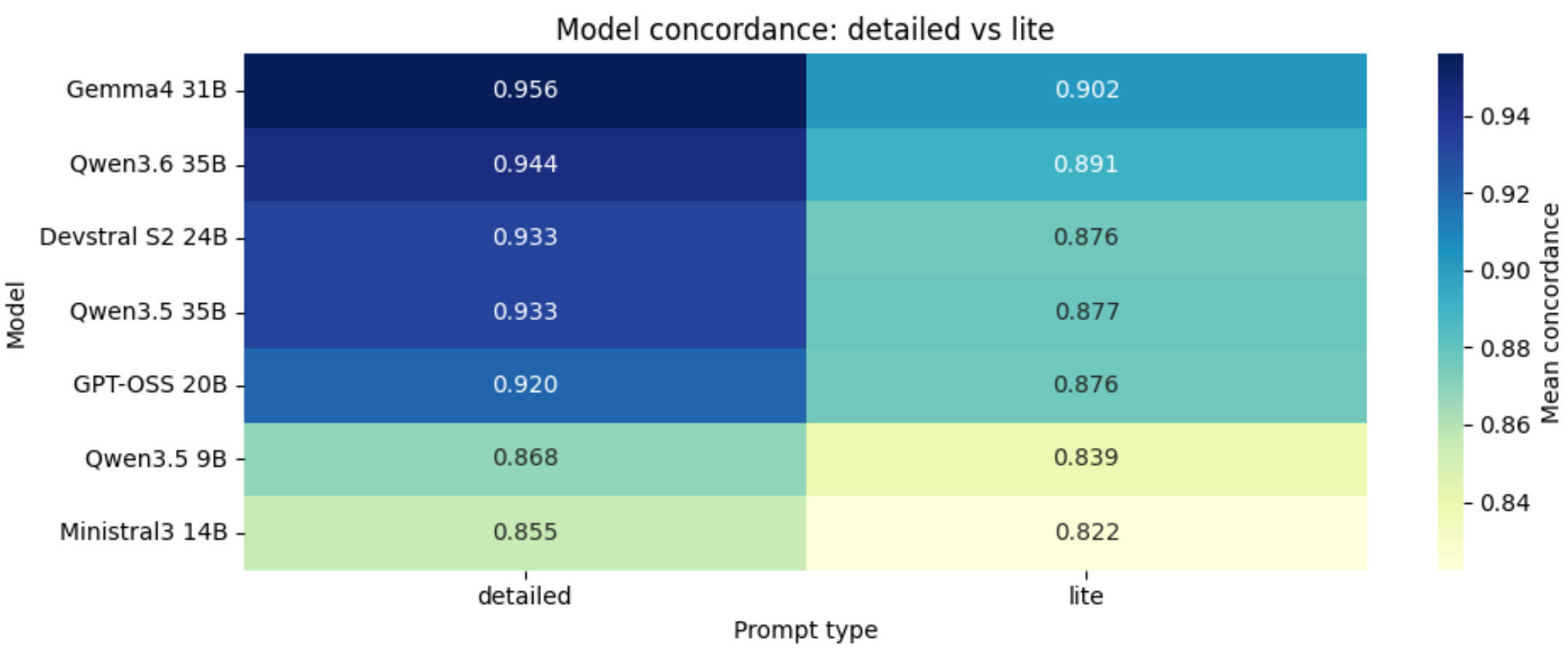

**Supplementary Figure 5:** Mean ground-truth vs predicted variable concordance.